\documentclass{article}
\usepackage{ijcai26}

\usepackage{times}
\usepackage{soul}
\usepackage{url}
\usepackage[hidelinks]{hyperref}
\usepackage[utf8]{inputenc}
\usepackage[small]{caption}
\usepackage{graphicx}
\usepackage{amsmath}
\usepackage{amsthm}
\usepackage{booktabs}
\usepackage{algorithm}
\usepackage{algorithmic}
\usepackage[switch]{lineno}
\usepackage{multirow}
\usepackage{amssymb}
\usepackage{threeparttable}

\title{From Values to Tokens: An LLM-Driven Framework for Context-Aware \\
Time Series Forecasting via Symbolic Discretization}

\author{
Xiaoyu Tao$^1$
\and
Shilong Zhang$^1$
\and
Mingyue Cheng$^1$\thanks{Corresponding Author.}
\and
Daoyu Wang$^1$
\and
Tingyue Pan$^1$
\and \\
Bokai Pan$^1$
\And
Changqing Zhang$^2$
\and
Shijin Wang$^3$\\
\affiliations
$^1$State Key Laboratory of Cognitive Intelligence, University of Science and Technology of China\\
$^2$College of Intelligence and Computing, Tianjin University\\
$^3$State Key Laboratory of Cognitive Intelligence, iFLYTEK Research\\
\emails
\{txytiny, zhangshilong, wdy030428, pty12345, bk2585934928\}@mail.ustc.edu.cn,
mycheng@ustc.edu.cn,
zhangchangqing@tju.edu.cn,
sjwang3@iflytek.com
}

\begin{document}

\maketitle

\begin{abstract}
Time series forecasting plays a vital role in supporting decision-making across a wide range of critical applications, including energy, healthcare, and finance. Despite recent advances,  forecasting accuracy remains limited due to the challenge of integrating historical numerical sequences with contextual features, which often comprise unstructured textual data. To address this challenge, we propose TokenCast, a large language model (LLM) driven framework that leverages language-based symbolic representations as a unified intermediary for context-aware time series forecasting. Specifically, TokenCast employs a discrete tokenizer to transform continuous numerical sequences into temporal tokens, enabling structural alignment with language-based inputs. To effectively bridge the semantic gap between modalities, both temporal and contextual tokens are embedded into a shared representation space via a pre-trained LLM, further optimized with generative objectives. Building upon this unified semantic space, the aligned LLM is subsequently fine-tuned in a supervised manner to predict future temporal tokens, which are then decoded back into the original numerical space. Extensive experiments on real-world datasets demonstrate the effectiveness of our framework and highlight its potential as a generative framework for context-aware time series forecasting. The code is available at \url{https://github.com/Xiaoyu-Tao/TokenCast}.
\end{abstract}

\section{Introduction}
Time series forecasting (TSF) is critical for decision-making in domains such as energy \cite{cheng2025comprehensive}, healthcare \cite{qiu2024tfb}, and finance \cite{feng2019temporal}. 
In practice, forecasting requires not only modeling temporal dependencies, but also understanding how they interact with external contextual factors—such as static attributes or dynamic events \cite{liu2024lstprompt}. 
Fundamentally, TSF can be viewed as learning a mapping from past values and contextual features to future outcomes \cite{jiang2025multi}. 

\begin{figure}
    \centering
    \includegraphics[width=1\linewidth]{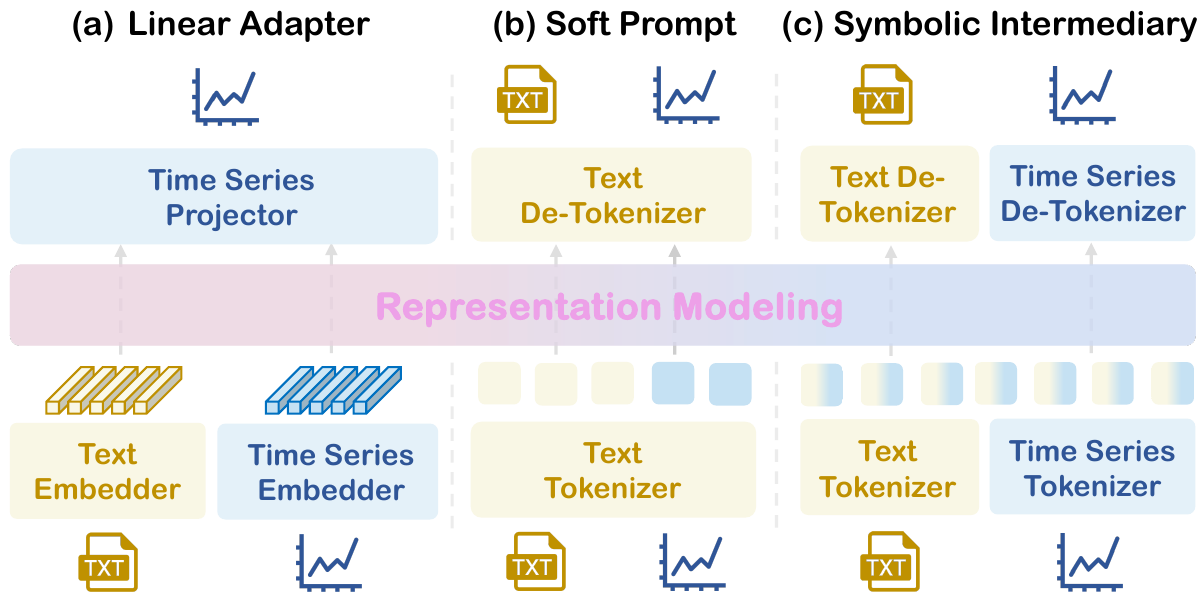}
    \caption{Methods for representation modeling of numerical sequences and contextual features: (a) linear adapter, (b) soft prompt, and (c) symbolic intermediary.}
    \label{fig:compared}
\end{figure}

To learn this mapping, researchers have proposed a comprehensive range of methods, ranging from classical statistical models to modern data-driven approaches. Traditional methods, such as ARIMA \cite{hyndman2008automatic} and state-space models \cite{winters1960forecasting}, rely on strong assumptions about data generation and often incorporate domain-specific priors. In contrast, recent data-driven approaches such as deep learning models aim to learn patterns directly from data without handcrafted assumptions. Architectures based on RNNs \cite{lai2018modeling}, CNNs \cite{cheng2025convtimenet},  Transformers \cite{zhou2022fedformer}, and MLPs \cite{zeng2023transformers} have been widely adopted, each capturing different aspects of temporal dependencies. However, most of these models assume homogeneous numerical inputs and struggle to effectively incorporate complex contextual features, particularly those with heterogeneous modalities.

Beyond capturing temporal dependencies, there is an increasing emphasis in recent research on incorporating contextual features to enhance forecasting performance \cite{williams2024context}.  These features typically fall into two categories: dynamic exogenous variables 
(e.g., weather conditions, event indicators) 
and static attributes (e.g., product types, patient demographics, market segments).
When contextual features share the same numerical modality as the target series, they can be directly modeled as additional input channels. However, many particularly high-value contextual features, such as clinical notes, policy texts, or user logs, are expressed in unstructured textual form. This heterogeneity poses significant challenges for aligning and integrating information across modalities.

To address these challenges, some studies have explored shallow fusion strategies to incorporate contextual features. Models such as DeepAR \cite{salinas2020deepar} and Temporal Fusion Transformer (TFT) \cite{lim2021temporal} typically concatenate external variables with time series or introduce gating mechanisms. While offering basic integration, these methods often rely on weak alignment and struggle to capture deep semantic interactions across modalities. More recently, LLMs have been introduced into time series forecasting \cite{sun2023test}. Methods like Time-LLM \cite{jin2023time} inject time series features into LLMs using linear adapters (Figure \ref{fig:compared} (a)) or soft prompts (Figure \ref{fig:compared} (b)). Although promising, these approaches fall short in resolving the structural discrepancies between numerical sequences and contextual features. Moreover, they fail to fully leverage the generative capabilities of LLMs, which are pretrained on large-scale corpora. This observation raises a fundamental question: \textit{Can time series be effectively modeled in a discrete token space to unlock the potential of LLMs}?

Motivated by this question, and as illustrated in Figure \ref{fig:compared}(c), we explore a more expressive paradigm that remains insufficiently explored in prior TSF literature, which formulates time series forecasting as a multimodal discrete context understanding and generation problem powered by pre-trained LLMs. The key idea is to transform continuous numerical sequences into discrete tokens and embed them into the same semantic space as contextual language inputs. This formulation enables the full use of LLMs’ capabilities in semantic understanding, contextual reasoning, and autoregressive generation. However, this paradigm introduces several non-trivial challenges. First, discretizing dynamic time series is more difficult than compressing static data, as it requires preserving temporal dependencies while reducing granularity. Second, even with symbolic representations, semantic misalignment between temporal tokens and contextual features may hinder effective fusion. Finally, it remains unclear whether time series forecasting can be effectively addressed through autoregressive generation over discrete tokens.

Based on the above analysis, we propose TokenCast, an LLM-driven framework for context-aware time series forecasting via symbolic discretization. TokenCast begins with a time series tokenizer that converts continuous sequences into discrete tokens, mitigating structural discrepancies across data modalities. To bridge the semantic gap, temporal and contextual tokens are jointly embedded into a shared representation space using a pre-trained LLM, optimized via a generative objective while keeping the backbone frozen and tuning only the embedding layer. Building on this unified semantic space, the aligned LLM is further fine-tuned with supervised forecasting signals to enhance predictive performance. We evaluate TokenCast on diverse real-world datasets enriched with contextual features. Experimental results show that TokenCast achieves strong accuracy and generalization across domains. We also conduct comprehensive ablation and qualitative studies, offering insights into the flexibility of symbolic, LLM-based time series forecasting.

\section{Related Work}

Time series forecasting (TSF) is a fundamental task across various domains. 
Traditional approaches typically rely on statistical assumptions such as stationarity and linearity, and often depend on handcrafted assumptions that limit their flexibility \cite{holt2004forecasting,kalekar2004time}.
Alternatively, data-driven methods \cite{chen2016xgboost}, particularly those based on deep learning, have advanced TSF by learning temporal patterns directly from data. RNN-based models \cite{wang2019deep} capture dependencies through recurrence, CNN-based models \cite{wang2023micn} enhance local pattern extraction, and Transformer-based architectures \cite{shi2024time} are well-suited for modeling long-range interactions. Furthermore, MLP-based approaches \cite{wang2024timemixer} demonstrate that simple architectures can achieve competitive performance with improved computational efficiency.
These models mainly focus on numerical data, with less emphasis on unstructured contextual features.

In addition to modeling temporal dependencies, recent research increasingly emphasizes the integration of contextual features for accurate forecasting \cite{chang2023llm4ts,hu2025context}. Two major lines of research have emerged in this direction. One line of research focuses on deep learning architectures that model feature interactions \cite{gasthaus2019probabilistic}. 
Another line of research leverages pre-trained LLMs for multimodal modeling \cite{cheng2025cross,cheng2025instructime}. 
Some approaches, such as TEMPO \cite{cao2023tempo}, utilize linear adapters to project temporal features into the LLM's semantic space. Others, like Promptcast \cite{xue2023promptcast}, employ soft prompts to guide the frozen LLM's behavior. 
However, these promising approaches fail to bridge the structural gap between numerical and textual modalities \cite{jia2024gpt4mts}.

\section{The Proposed TokenCast}
In this section, we present the formal problem definition, clarify the key concepts and notations used consistently throughout the paper, and provide an overview of the TokenCast. 

\begin{figure*}[t!]
    \centering
    \includegraphics[width=1\linewidth]{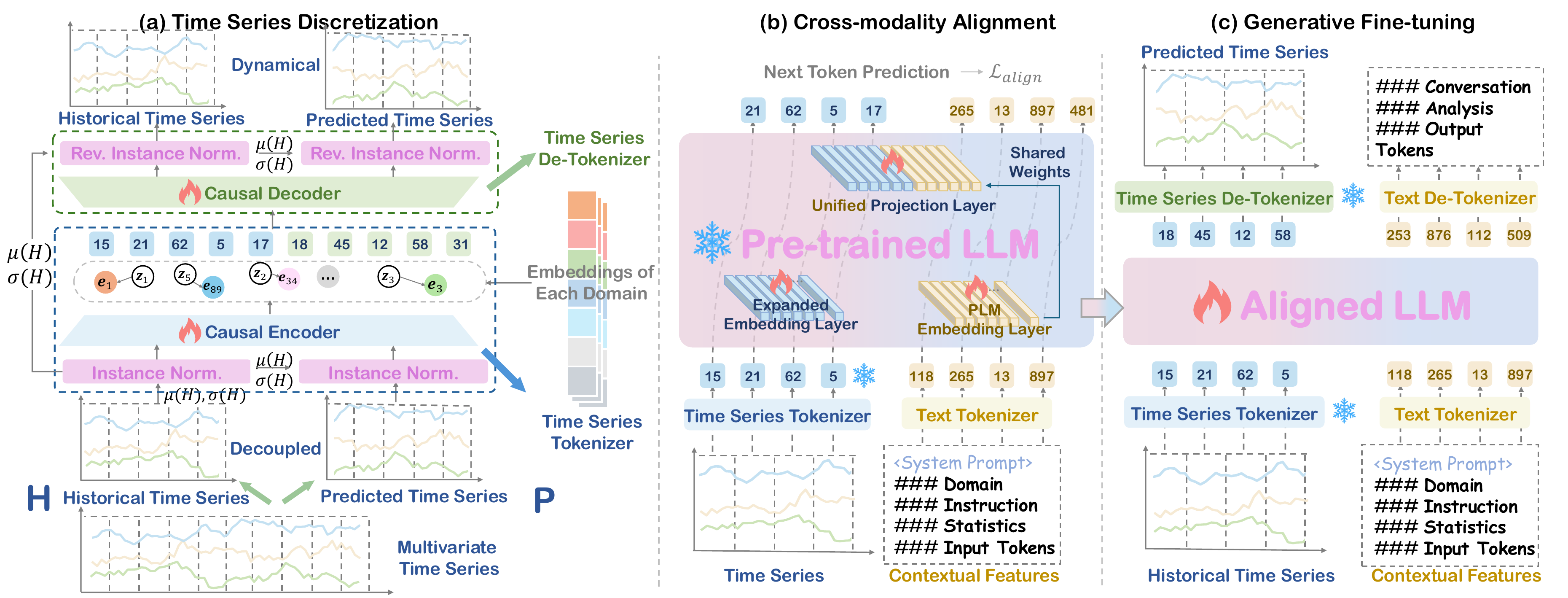}
    \caption{Overview of the framework for context-aware time series forecasting: (a) time series tokenizer to address the structural differences between modalities, (b) cross-modality alignment with a generative objective to bridge the modalities, and (c) generative fine-tuning and context-aware forecasting through time series decoding for horizon prediction.}
    \label{fig:main}
\end{figure*}
\subsection{Problem Formulation}

We consider a dataset $\mathcal{D} = \{(H_i, T_i, P_i)\}_{i=1}^{N}$ of $N$ multimodal instances, where $H \in \mathbb{R}^{L \times C}$ denotes the historical multivariate time series, $T$ denotes the contextual features, and $P \in \mathbb{R}^{L_P \times C}$ denotes the future time series. The contextual features $T$ are tokenized into language tokens $Y$ by a pre-trained LLM tokenizer, while $H$ is mapped to discrete time series tokens $Z_q$ through a learnable function $f_\theta: H \mapsto Z_q$. These tokens are concatenated as $Z = [Z_q; Y] \in \mathcal{V}^{T'}$. With boundary markers delimiting the generated temporal tokens $\hat{Z}$, a decoding function $g_\phi : \hat{Z} \mapsto \hat{P}$ reconstructs the predicted time series $\hat{P} \in \mathbb{R}^{L_P \times C}$.
\subsection{Framework Overview}
Figure \ref{fig:main} provides an overview of the TokenCast, which consists of three main stages. The process begins with the  discretization, which transforms continuous sequences into a sequence of discrete tokens via a decoupled and dynamical vector quantization tokenizer. Subsequently, both the temporal and contextual tokens are then jointly processed by a pre-trained LLM, which performs cross-modality alignment under generative objectives. Following this alignment, the aligned LLM is adapted to the downstream task via generative fine-tuning. The predicted tokens are decoded to raw time series using a frozen de-tokenizer. The following sections elaborate on the principal stages of the TokenCast.

\subsection{Time Series Discretization}
\paragraph{Time Series Tokenizer.} To fully harness the generative and reasoning capabilities of language models, symbolic representation naturally arises as an effective intermediary. Accordingly, we employ time series discretization as a simple yet powerful approach to establish this bridge. 
It is worth noting that existing approaches, such as Symbolic Aggregate Approximation (SAX) \cite{lin2007experiencing}, have achieved progress in time series discretization but often suffer from significant information loss due to dimensionality reduction. In contrast, reconstruction-based methods map subsequences to discrete codes from a predefined codebook and achieve more precise representations through reconstruction optimization.
While preserving the original information is advantageous, previous reconstruction-based methods typically encode the entire sequence, overlooking the statistical properties of time series. 
Reversible Instance Normalization (RevIN) \cite{kim2021reversible} is widely used in forecasting, but its reliance on cached normalization statistics can lead to future information leakage when applied over prediction horizons. To mitigate this issue, we introduce a decoupled and dynamic tokenizer.

As illustrated in Figure \ref{fig:main} (a), similar to the forecasting phase, we divide the multivariate time series into a historical time series $H \in \mathbb{R}^{L_H \times C}$ and a predicted time series $P \in \mathbb{R}^{L_P \times C}$, which can be formally represented as $X = [H; P] \in \mathbb{R}^{L \times C}$.
The process begins with a reversible instance normalization (RIN) layer. We compute the mean $\mu(H)$ and standard deviation $\sigma(H)$ solely from the historical time series $H$, and apply them to normalize the time series $X$, thereby preventing future information leakage. These statistics are retained for inverse transformation during decoding. Instead of employing separate encoders, we adopt a shared encoder, which facilitates the joint modeling of both local and global information.
The normalized time series is then passed through a causal encoder $f_{\text{enc}}$, yielding a sequence of continuous latent representations $Z = f_{\text{enc}}(X) \in \mathbb{R}^{T \times d}$, where $T$ is the number of latent vectors and $d$ is the feature dimension. To discretize the latent representations, we apply a vector quantization layer. For domain $i$, a learnable codebook $C_i = \{ e_{i,k} \}_{k=1}^{K} \subset \mathbb{R}^d$ is maintained, containing $K$ embedding vectors. Each latent vector $z_t \in \mathbb{R}^d$ is mapped to its nearest neighbor in the codebook as $z_t^q = e_{i,k^*}$, where $k^* = \arg\min_k \| z_t - e_{i,k} \|_2^2$. The output of this layer is a quantized sequence $Z_q = (z_1^q, \dots, z_T^q)$, and the corresponding sequence of indices $\{k^*\}$ serves as the discrete tokens for downstream modeling. 
These tokens are subsequently decoded by a shared causal decoder $f_{\text{dec}}$, rather than by separate decoders, which ensures consistent reconstruction and enables the predicted part to dynamically exploit richer historical features. Then, the final reconstruction $\hat{X}$ is obtained by applying the inverse RIN operation using the stored statistics $\mu(H)$ and $\sigma(H)$, i.e., $\hat{X} = f_{\text{denorm}}(f_{\text{dec}}(Z_q))$.

\paragraph{Training Objective.} The tokenizer is optimized by minimizing the objective function defined as follows:
\begin{equation}
\mathcal{L} = \mathcal{L}_{\text{recon}} + \beta \left( \mathcal{L}_{\text{commit}} + \mathcal{L}_{\text{codebook}} \right) + \gamma \mathcal{L}_{\text{diversity}},
\end{equation}
where $\mathcal{L}_{\text{recon}} = \| \hat{X} - X \|_2^2$ is the reconstruction loss that optimizes both the encoder and decoder. Due to the non-differentiability of the $\arg\min$ operation in quantization, we employ the straight-through estimator (STE) during backpropagation.
To train the vector quantizer, we include:
$\mathcal{L}_{\text{codebook}} = \| \text{sg}[Z] - Z_q \|_2^2$,  
$\mathcal{L}_{\text{commit}} = \| Z - \text{sg}[Z_q] \|_2^2$,
where $\text{sg}[\cdot]$ denotes the stop-gradient operator, which prevents gradients from flowing into its argument during backpropagation.
To promote diverse usage of codebook entries, we further add a diversity loss
$\mathcal{L}_{\text{diversity}} = \tfrac{1}{N} \sum_{i=1}^N \tfrac{1}{d_i + \epsilon}$,  
where $d_i = \min_{j \neq i}\| e_i - e_j \|_2$ denotes the nearest-neighbor distance 
between codebook embeddings. This penalty discourages vectors from clustering too closely 
and encourages more uniform utilization of the codebook.

\subsection{Pre-trained LLM Backbone Formulation}

Following the discretization of time series into discrete tokens, the next challenge is to model the complex dependencies embedded in these sequences. While architectures like TCNs or Transformers can be trained from scratch, we argue that a pre-trained LLM serves as a more effective backbone.  This is supported by two observations: 
(1) a pre-trained LLM possesses strong semantic understanding and contextual reasoning capabilities acquired from large-scale corpora, and (2) the structure of discrete time series tokens closely resembles that of language tokens \cite{zhao2023survey}. By casting forecasting as a generative task, we directly leverage the LLM’s autoregressive generation ability.
To guide LLM reasoning and incorporate contextual features, we employ a structured prompt template, as shown in Figure \ref{fig:main} (b). This prompt template consists of four essential components: domain knowledge, task instructions, statistical properties, and discrete time series tokens. This design ensures token-level consistency with language tokens and introduces task-specific descriptions alongside statistical attributes, enabling the LLM to perform instruction-driven generation. 

\subsection{Cross-Modality Alignment of Time Series and Contextual Features}
While discretization aligns time series structurally with language tokens, a semantic gap remains between time series and contextual features. Existing methods often introduce projection modules (e.g., MLPs) to map time series into the LLM’s latent space for fusion \cite{jia2024gpt4mts}. Although effective in downstream tasks, these strategies rely on external transformation modules for alignment, which bypass the language model’s native vocabulary modeling mechanism.
To this end, we implement a more explicit vocabulary-level alignment strategy. As illustrated in Figure \ref{fig:main} (b), we construct a unified vocabulary by directly appending $K$  temporal tokens and $S$ task-specific special tokens to the original vocabulary $V_{\text{orig}}$ of the pre-trained LLM, forming an extended vocabulary $V$. 
Correspondingly, a shared embedding matrix $E \in \mathbb{R}^{|V| \times d}$ is used to encode all tokens, regardless of their modality origin. 
This unified embedding mechanism enables seamless fusion of time series and contextual features while maintaining alignment with the pre-trained model.
To ensure distributional alignment with pretrained embeddings for fine-tuning, the embedding of the newly introduced time series tokens is initialized by sampling from a multivariate gaussian distribution defined by the mean $\mu$ and covariance $\Sigma$ of the original word embeddings. Then, temporal tokens $Z_q$ and contextual tokens $Y$ are concatenated at the token level and jointly transformed into embeddings via the shared embedding layer:
\(E([Z_q, Y]) = [E(z_1), \dots, E(z_n), E(y_1), \dots, E(y_m)]\),
where $E$ denotes the unified embedding matrix. This unified embedding process enables the LLM to reason over concatenated sequences without requiring architectural modification. 

To optimize cross-modality token representations within the shared embedding space, we adopt an autoregressive training objective. Specifically, we freeze all parameters of the pre-trained LLM and update only the shared embedding matrix \(E\), which is responsible for encoding both temporal and contextual tokens. Given a concatenated token sequence \([Z_q, Y]\), the training objective is formulated as a next-token prediction task over the combined sequence:
\begin{equation}
\mathcal{L}_{\text{align}} = -\sum_{t=1}^{T} \log p(z_t \mid z_1, \dots, z_{t-1}; E),
\end{equation}
where \(z_t \in V\) denotes the $t$-th token in the sequence, and \(p(\cdot)\) is the conditional probability predicted by the frozen language model given the embedding vectors from \(E\).

\subsection{Generative Fine-tuning and Context-aware Time Series Forecasting}

We now detail the procedure for adapting the aligned LLM for forecasting tasks. As illustrated in Figure \ref{fig:main} (c), we employ a generative fine-tuning strategy to specialize the model for context-aware time series forecasting. This process consists of two primary stages: (1) structured prompt-based generative fine-tuning; and (2) context-aware time series forecasting with token-based decoding.

In the first stage, prompt-based generative fine-tuning is introduced to explicitly transfer the pretrained language modeling capability into the forecasting domain. Instead of relying on external mapping modules, generative fine-tuning directly formulates forecasting as a generation task, where the model is supervised to output both natural language reasoning and sequences of future time series tokens. This paradigm fosters a fast-thinking behavior: by optimizing an autoregressive objective against ground-truth structured responses, the model learns to rapidly recognize patterns, associate contextual features with temporal dynamics, and produce coherent outputs without engaging in deep deliberation. As a result, the aligned LLM acquires the ability to generate fluent and context-aware predictions.
In the second stage, the fine-tuned model is utilized for context-aware forecasting and decoding. During inference, the model receives a prompt with historical data and contextual features, and autoregressively generates a complete response. The key component of this generated output is the sequence of discrete tokens, which represents the model's prediction of future time series values. To translate this symbolic representation back into a continuous predicted time series, these tokens are processed by a frozen time series de-tokenizer. We use boundary markers to delimit the temporal tokens.

\begin{table*}[t]
    \setlength{\tabcolsep}{3.5pt} 
    \small
    \centering
    \begin{threeparttable}
    \resizebox{\textwidth}{!}{%
    \begin{tabular}{c|c|cc|cccc|cccc|cccccc}
    \toprule
    \multicolumn{2}{c|}{Model}      & \multicolumn{2}{c|}{\textbf{TokenCast}} & \multicolumn{2}{c}{\textbf{Time-LLM}} & \multicolumn{2}{c|}{\textbf{GPT4TS}} & \multicolumn{2}{c}{\textbf{TimeDART}} & \multicolumn{2}{c|}{\textbf{SimMTM}} & \multicolumn{2}{c}{\textbf{Crossformer}} & \multicolumn{2}{c}{\textbf{Autoformer}} & \multicolumn{2}{c}{\textbf{DLinear}} \\
    \multicolumn{2}{c|}{Metric} & MSE & MAE & MSE & MAE & MSE & MAE & MSE & MAE & MSE & MAE & MSE & MAE & MSE & MAE & MSE & MAE \\
    \midrule

    \multicolumn{2}{c|}{Economic}
    & \textbf{68.911} & \textbf{1.701} & \underline{81.542} & 1.760 & 85.947 & \underline{1.716} & 86.029 & 1.771 & 90.351 & 1.672 & 406.418 & 4.074 & 116.745 & 2.088 & 122.216 & 2.070 \\
    \cmidrule{1-18}

    \multicolumn{2}{c|}{Health}
    & \textbf{2.525} & \textbf{0.081} & 2.823 & 0.104 & \underline{2.565} & \underline{0.083} & 2.623 & 0.088 & 2.720 & 0.088 & 1644.745 & 2.504 & 2.617 & 0.265 & 28.587 & 0.455 \\
    \cmidrule{1-18}

    \multicolumn{2}{c|}{Web}
    & \textbf{497.410} & \textbf{1.246} & 557.833 & 1.751 & \underline{540.492} & 1.458 & 773.635 & 1.369 & 847.649 & \underline{1.327} & 698.316 & 1.963 & 722.506 & 3.303 & 632.301 & 1.398 \\
    \cmidrule{1-18}

    \multicolumn{2}{c|}{Stock-NY}
    & \textbf{0.482} & \textbf{0.455} & 0.662 & 0.510 & 0.638 & \underline{0.502} & 0.776 & 0.606 & \underline{0.613} & 0.585 & 1.111 & 0.912 & 0.676 & 0.573 & 0.999 & 0.754 \\
    \cmidrule{1-18}

    \multicolumn{2}{c|}{Stock-NA}
    & \textbf{1.134} & \textbf{0.780} & \underline{1.200} & 0.925 & 1.272 & 0.880 & 1.409 & 0.883 & 1.343 & \underline{0.834} & 1.913 & 1.053 & 1.558 & 0.914 & 1.710 & 0.958 \\
    \cmidrule{1-18}

    \multicolumn{2}{c|}{Nature}
    & 0.269 & 0.297 & \underline{0.258} & \underline{0.283} & 0.274 & 0.299 & \textbf{0.243} & \textbf{0.273} & 0.259 & 0.286 & 0.735 & 0.511 & 0.508 & 0.481 & 0.369 & 0.436 \\
    \cmidrule{1-18}

    \multicolumn{2}{c|}{1\textsuperscript{st} Count} & \textbf{5} & \textbf{5} & 0 & 0 & 0 & 0 & 1 & 1 & 0 & 0 & 0 & 0 & 0 & 0 & 0 & 0 \\

    \bottomrule
    \end{tabular}
    }
    \end{threeparttable}
    \caption{All reported results are \textbf{average} over four horizons and three trials on various context-rich benchmark datasets. Lower values indicate better performance. The best results are highlighted in \textbf{bold}, and the second-best are \underline{underlined}.}
    \label{tab:avg_main_results}
\end{table*}
\section{Experiments}

In this section, we conduct comprehensive experiments to evaluate our TokenCast's performance on diverse real-world datasets enriched with contextual features for time series forecasting. Additionally, we perform extensive ablation studies and exploration analysis to demonstrate the effectiveness of its individual components.
\begin{table}[t]
  \centering
  \resizebox{\columnwidth}{!}{%
  \begin{tabular}{ccccc}
    \toprule
    \textbf{Dataset} & \textbf{Domain} & \textbf{Frequency} & \textbf{Length} & \textbf{Variables} \\
    \midrule
    
    Economic & Economic & 1 month & 728 & 107 \\
    Health & Health & 1 day & 1,392 & 948 \\
    Web & Web & 1 day & 792 & 2,000 \\
    Stock-NY & Stock & 1 day & 1,243 & 5 \\
    Stock-NA & Stock & 1 day & 1,244 & 5 \\
    Nature & Nature & 30 mins & 19,934 & 11 \\
    \bottomrule
  \end{tabular}
  }
  \caption{Diverse real-world datasets from various domains and with distinct characteristics. }
  \label{tab:dataset}
\end{table}

\subsection{Experimental Settings}

\paragraph{Datasets.}
As shown in Table \ref{tab:dataset}, we evaluate our framework on six real-world datasets from diverse domains enriched with contextual features:  \textbf{Economic} \cite{mccracken2016fred}, \textbf{Health} \cite{panagopoulos2021transfer}, \textbf{Web} \cite{gasthaus2019probabilistic}, two subsets of \textbf{Stock} data \cite{feng2019temporal} and \textbf{Nature} \cite{poyatos2020global}. These datasets, spanning various temporal patterns and contextual dependencies, serve as a comprehensive benchmark for context-aware forecasting. Data preparation involves imputing missing values and applying z-score normalization to all datasets, ensuring stable convergence and comparability. 
\paragraph{Baselines.}
We compare our proposed framework against eight strong baselines, grouped into four representative categories for comprehensive evaluation. For LLM-based models, we include Time-LLM \cite{jin2023time} and GPT4TS \cite{zhou2023one}, which adapt pre-trained LLMs for time series forecasting using modality-aware prompting and reprogramming. In the self-supervised frameworks category, we evaluate TimeDART \cite{wang2024timedart} and SimMTM \cite{dong2023simmtm}.
Additionally, we include Transformer-based methods like Autoformer \cite{wu2021autoformer} and Crossformer \cite{zhang2023crossformer}. Finally, we consider the MLP-based method DLinear \cite{zeng2023transformers}. Further details are provided in the Appendix.

\paragraph{Implementation Details.}  For each baseline, we search over multiple input lengths and report the best performance to avoid underestimating its capability. The historical length is set to $L = 96$ for the Nature dataset and $L = 36$ for the other five datasets, based on the data volume and temporal resolution. The forecasting horizons are set to \{24, 48, 96, 192\} for Nature and \{24, 36, 48, 60\} for the other dataset. We adopt two widely used evaluation metrics in time series forecasting: mean absolute error (MAE) and mean squared error (MSE). We report average results for the main and ablation studies. For exploratory analysis, we use 96-to-24 on Nature and 36-to-24 on the other datasets. Complete results for the main experiments, ablation studies, and exploratory analysis are included in the Appendix. All experiments are implemented in PyTorch and conducted on a distributed setup with 8 NVIDIA A100 GPUs. 


\begin{table*}[ht]
\small
\setlength{\tabcolsep}{3.5pt}
\centering
\resizebox{\textwidth}{!}{%
\begin{tabular}{c|cc|cc|cc|cc|cc|cc} 
\toprule

\multirow{2}{*}{\textbf{Model Variants}} & 
\multicolumn{2}{c|}{\textbf{Economic}} & 
\multicolumn{2}{c|}{\textbf{Health}} & 
\multicolumn{2}{c|}{\textbf{web}} & 
\multicolumn{2}{c|}{\textbf{Stock-NY}} & 
\multicolumn{2}{c|}{\textbf{Stock-NA}} & 
\multicolumn{2}{c}{\textbf{Nature}} \\

\cmidrule(lr){2-3} \cmidrule(lr){4-5} \cmidrule(lr){6-7} \cmidrule(lr){8-9} \cmidrule(lr){10-11} \cmidrule(lr){12-13}

& MSE & MAE & MSE & MAE & MSE & MAE & MSE & MAE & MSE & MAE & MSE & MAE \\
\midrule
w/ Cross-Modality Alignment, w/o Generative Fine-Tuning & 80.418 & 1.774 & 2.875 & 0.084 & 555.375 & 1.447 & 0.556 & 0.479 & 1.317 & 0.813 & 0.378 & 0.357 \\
w/o Cross-Modality Alignment, w/ Generative Fine-Tuning & 72.292 & 1.690 & 2.783 & 0.079 & 504.740 & 1.264 & 0.515 & 0.478 & 1.181 & 0.804 & 0.305 & 0.318 \\
\textbf{w/ Cross-Modality Alignment, w/ Generative Fine-Tuning} & \textbf{68.911} & \textbf{1.701} & \textbf{2.524} & \textbf{0.081} & \textbf{497.410} & \textbf{1.246} & \textbf{0.482} & \textbf{0.455} & \textbf{1.134} & \textbf{0.780} & \textbf{0.269} & \textbf{0.297} \\
\bottomrule
\end{tabular}
}
\caption{Ablation study on the effects of cross-modality alignment and generative fine-tuning across multiple datasets.}
\label{tab:ablation_pretrain_sft}
\end{table*}
\subsection{Forecasting Performance Analysis}
Table~\ref{tab:avg_main_results} comprehensively compares forecasting performance across six benchmark datasets. TokenCast demonstrates superior performance in most scenarios, further confirming previous empirical findings \cite{zhou2023one} that no single model performs best across all settings. 
Notably, LLM-based baselines like Time-LLM also show competitive results, particularly on context-rich datasets such as Economic and Stock-NY. This further validates the potential of leveraging large language models in time series forecasting. However, these models often lack the structural alignment mechanisms introduced by our framework, limiting their consistent performance.
Conventional baselines such as TimeDART perform well on datasets with strong periodicity and weak contextual dependence (e.g., Nature), but their performance drops significantly on complex datasets rich in contextual features (e.g., Economic and Web). This contrast underscores the importance of contextual feature modeling and cross-modal interaction.
In summary, our framework delivers state-of-the-art results with high consistency. This is attributed to its core design: discretizing time series into discrete tokens and aligning them with contextual features. This unified token-based paradigm effectively addresses real-world context-aware time series forecasting challenges.

\subsection{Ablation Studies}
\paragraph{Ablation on Alignment and Fine-tuning.}
We conduct the ablation study on two crucial training steps: the cross-modality alignment and generative fine-tuning. The comprehensive results in Table \ref{tab:ablation_pretrain_sft} clearly demonstrate their indispensable contribution to the overall framework performance. The model equipped with the cross-modality alignment stage consistently achieves lower MSE scores across all six datasets. Without this alignment, contextual features risk being misinterpreted by the time series backbone, leading to suboptimal forecasts. This highlights its critical role in effectively integrating contextual information by bridging structural and semantic discrepancies between time series and contextual features, thus facilitating meaningful feature interaction. This alignment thus acts as a foundational step, ensuring the subsequent fine-tuning stage operates on a semantically rich and coherent feature space.

Concurrently, Table \ref{tab:ablation_pretrain_sft} vividly illustrates the pivotal contribution of the generative fine-tuning stage. Across all six benchmark datasets, the model employing generative fine-tuning consistently and substantially outperforms its counterpart that omits this crucial step. The performance degradation when omitting this stage is notable across various datasets, underscoring the general applicability and importance of the fine-tuning process. This drop is particularly stark on datasets like Stock-NA, where the complex, non-stationary patterns demand task-specific adaptation. Ultimately, these findings emphasize that generative fine-tuning is essential for adapting the pre-trained LLM's general capabilities to generative time series forecasting.

\begin{figure}[t]
    \centering
    \includegraphics[width=1\linewidth]{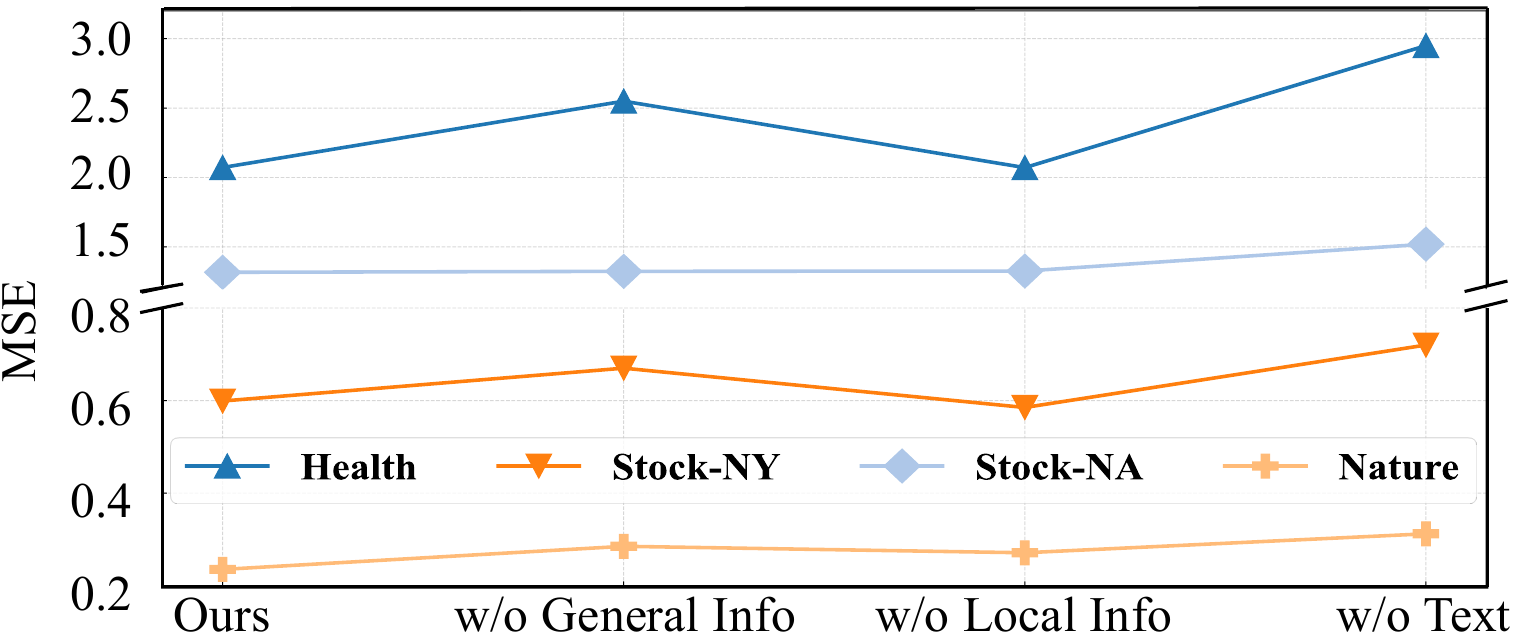}
    \caption{Ablation study on multiple datasets on the contribution of multimodal context in time series forecasting. }
    \label{fig:text}
\end{figure}

\paragraph{Ablation on Multimodal Contributions.}
Fig. ~\ref{fig:text} analyzes how different types of contextual features affect forecasting performance. Incorporating any contextual features yields substantial improvements, as the variant without contextual input consistently performs worse. We further divide the input into general info, which provides high-level context such as domain knowledge and task instructions, and local info, which offers event-specific details like static statistical attributes. While both types contribute to performance, general info typically brings larger improvements. The strong results of the full model indicate its ability to effectively combine the broad context from general info and the specific cues from local info to enhance prediction accuracy.

\begin{table}[t!]
\small
\setlength{\tabcolsep}{3.5pt}
 \centering
 \resizebox{\columnwidth}{!}{%
 \begin{tabular}{c|ccc|ccc|ccc}
 \toprule
 \multicolumn{1}{c|}{Dataset} & \multicolumn{3}{c|}{\textbf{Economic}} & \multicolumn{3}{c|}{\textbf{Stock-NA}} & \multicolumn{3}{c}{\textbf{Nature}} \\
 \multicolumn{1}{c|}{Metrics} & Recon. & MSE & MAE & Recon. & MSE & MAE & Recon. & MSE & MAE \\
 \midrule
 32 & 190.371 & 50.234 & 1.372 & 0.244 & 0.794 & 0.636 & 0.134 & 0.233 & 0.281 \\
 64 & \textbf{141.852} & \textbf{37.699} & \textbf{1.293} & 0.213 & 0.690 & 0.616 & 0.158 & 0.241 & 0.296 \\
128 & 170.630 & 39.379 & 1.251 & \textbf{0.205} & \textbf{0.571} & \textbf{0.600} & \textbf{0.104} & \textbf{0.203} & \textbf{0.265} \\
 256 & 191.937 & 39.309 & 1.339 & 0.209 & 0.646 & 0.593 & 0.114 & 0.248 & 0.288 \\
 \bottomrule
 \end{tabular}
 }
 \caption{Study on the number of tokens in the codebook across multiple datasets. We report predicted reconstructed MSE (Recon.), downstream MSE, and downstream MAE.}
 \label{tab:ablation_num_tokens_select}
\end{table}

\subsection{Exploration Analysis}

\paragraph{Codebook Size.}
We conduct a study to assess the impact of codebook size on model performance, as summarized in Table~\ref{tab:ablation_num_tokens_select}.
The results highlight the importance of selecting an appropriate codebook size for time series forecasting.
Specifically, a size of 128 achieves state-of-the-art results on the Nature and Stock-NA datasets, while a smaller size of 64 excels on the Economic dataset. Interestingly, both smaller (32) and larger (256) codebook sizes fail to produce better results and often lead to significant performance degradation. 
This suggests that for our framework, simply increasing token granularity is not always beneficial. Instead, a moderate codebook size strikes a balance between reconstruction fidelity and the complexity of the downstream task.

\begin{figure}[b]
    \centering
    \includegraphics[width=1\linewidth]{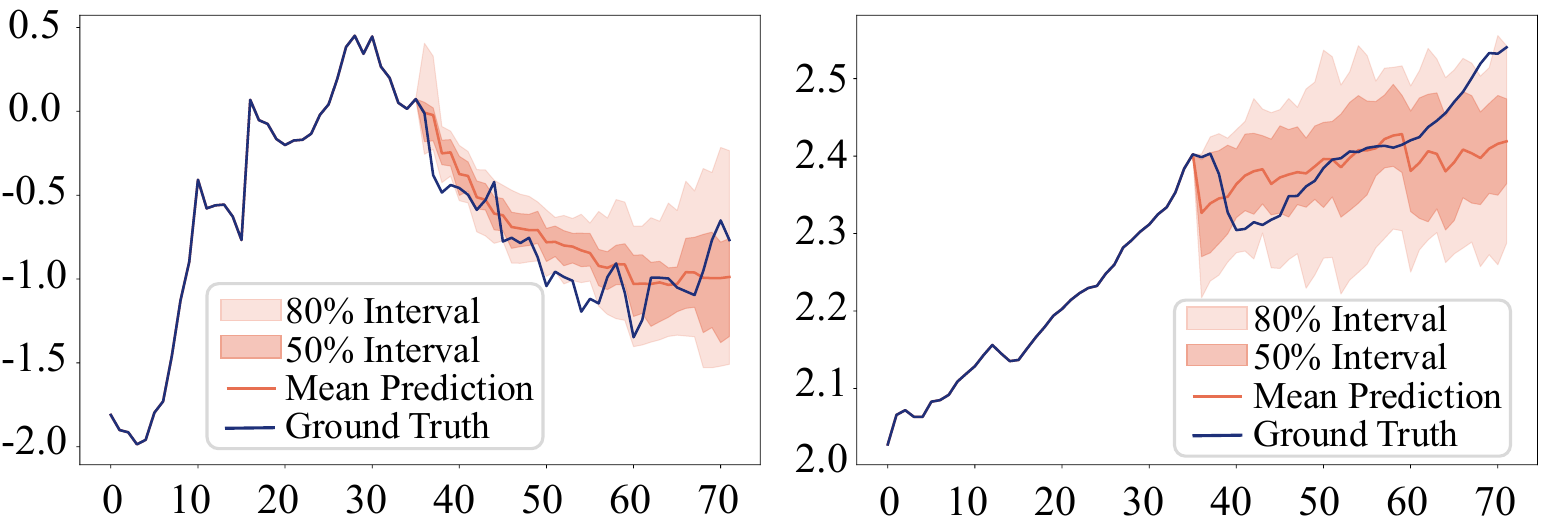}
    \caption{Forecasting with uncertainty on Stock-NY (left) and Economic (right) datasets. The plots compare the ground truth trajectories with the model’s mean predictions, along with the 50\% and 80\% predictive intervals.}
    \label{fig:uncertainty}
\end{figure}

\paragraph{Generative Uncertainty.}

To validate the uncertainty modeling capabilities of our TokenCast, we conduct experiments on both the Economic and Stock-NY datasets. As shown in Fig.~\ref{fig:uncertainty}, our method produces predictive distributions that closely track the ground truth, with 50\% and 80\% prediction intervals capturing the inherent variability in the data. By adjusting the temperature during sampling, we observe that the model can flexibly modulate the spread of the predictive intervals, indicating its potential for controllable uncertainty-aware forecasting. This demonstrates that our model not only provides accurate mean predictions but also yields well-calibrated uncertainty estimates.

\paragraph{LLM Backbone.}
We evaluate four LLM backbones to identify the optimal architecture for our forecasting framework. As summarized in Table~\ref{tab:backbone_comparison}, the Qwen2.5-0.5B-base models consistently demonstrate superior performance. Specifically, the base version achieves state-of-the-art results on the Nature and Stock-NA datasets, while the instruct-tuned version excels on the more complex Economic dataset.
Interestingly, larger models like Qwen2.5-1.5B-inst. fail to yield further gains and often underperform. This suggests that for our tasks, simply scaling up model size is not beneficial. Instead, the 0.5B models strike a balance between representational capacity and generalization.

\begin{table}[t!]
    \centering
    \small
    \resizebox{\columnwidth}{!}{%
    \begin{tabular}{c | cc | cc | cc}
    \toprule
    \multicolumn{1}{c|}{Dataset} & \multicolumn{2}{c|}{\textbf{Economic}} & \multicolumn{2}{c|}{\textbf{Stock-NA}} & \multicolumn{2}{c}{\textbf{Nature}} \\
    \multicolumn{1}{c|}{Metrics} & MSE & MAE & MSE & MAE & MSE & MAE \\
    \midrule
    Qwen2.5-0.5B-base    & 37.164         & 1.301          & \textbf{0.668} & \textbf{0.605} & \textbf{0.180} & \textbf{0.246} \\
    Qwen2.5-0.5B-inst.   & \textbf{36.744} & \textbf{1.299} & 0.695          & 0.614          & 0.187          & 0.253 \\
    Qwen2.5-1.5B-inst.   & 38.549         & 1.283          & 0.722          & 0.611          & 0.229          & 0.270 \\
    Qwen3-0.6B-inst.     & 39.629         & 1.315          & 0.936          & 0.715          & 0.236          & 0.281 \\
    \bottomrule
    \end{tabular}
    }
    \caption{Performance comparison of different backbones and their variants (base/instruct) across varying model scales.}
    \label{tab:backbone_comparison}
\end{table}

\paragraph{Embedding Layer Initialization.}
We investigate three initialization strategies for our model's embedding layer to identify the most effective approach. As shown in Table \ref{tab:init_comparison}, mean initialization consistently provides the most consistent performance. Specifically, it achieves the best results on the Nature and Economic datasets. While word initialization is superior on the Stock-NA dataset, its performance is less consistent across other domains. Notably, standard random initialization suffers a significant performance degradation on Stock-NA, highlighting its instability. These findings suggest that initializing embeddings with meaningful prior information provides a better starting point for optimization. Therefore, we adopt mean initialization as the default.

\begin{table}[t!]
    \centering
    \small
    \resizebox{\columnwidth}{!}{%
    \begin{tabular}{c | cc | cc | cc}
    \toprule
    \multicolumn{1}{c|}{Dataset} & \multicolumn{2}{c|}{\textbf{Economic}} & \multicolumn{2}{c|}{\textbf{Stock-NA}} & \multicolumn{2}{c}{\textbf{Nature}} \\
    \multicolumn{1}{c|}{Metrics} & MSE & MAE & MSE & MAE & MSE & MAE \\
    \midrule
    Mean Initialization         & \textbf{36.744} & \textbf{1.299} & 0.695          & 0.614          & \textbf{0.187} & \textbf{0.253} \\
    Word Initialization          & 39.680          & 1.261          & \textbf{0.667} & \textbf{0.602} & 0.224          & 0.264          \\
    Random Initialization       & 36.744          & 1.299          & 1.101          & 0.725          & 0.189          & 0.256          \\
    \bottomrule
    \end{tabular}
    }
    \caption{Study on different initialization methods on the embedding layer. We compare mean initialization, word initialization, and random initialization.}
    \label{tab:init_comparison}
\end{table}

\begin{figure}[t!]
    \centering
    \includegraphics[width=1\linewidth]{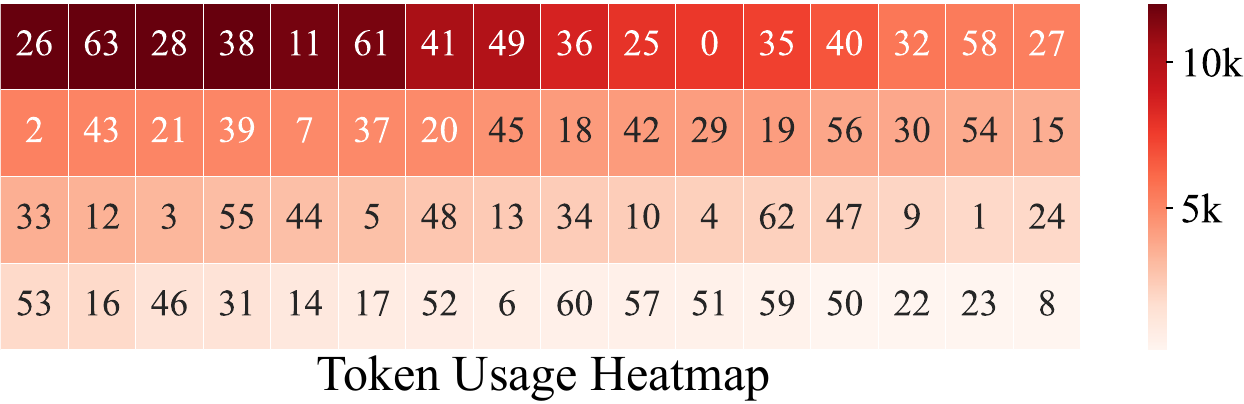}
    \caption{Token usage statistics over the Nature codebook.
The heatmap shows the usage frequency of all 64 tokens, with color intensity reflecting how often each token appears.}
    \label{fig:codebook_heatmap}
\end{figure}

\begin{figure}[t!]
    \centering
    \includegraphics[width=1\linewidth]{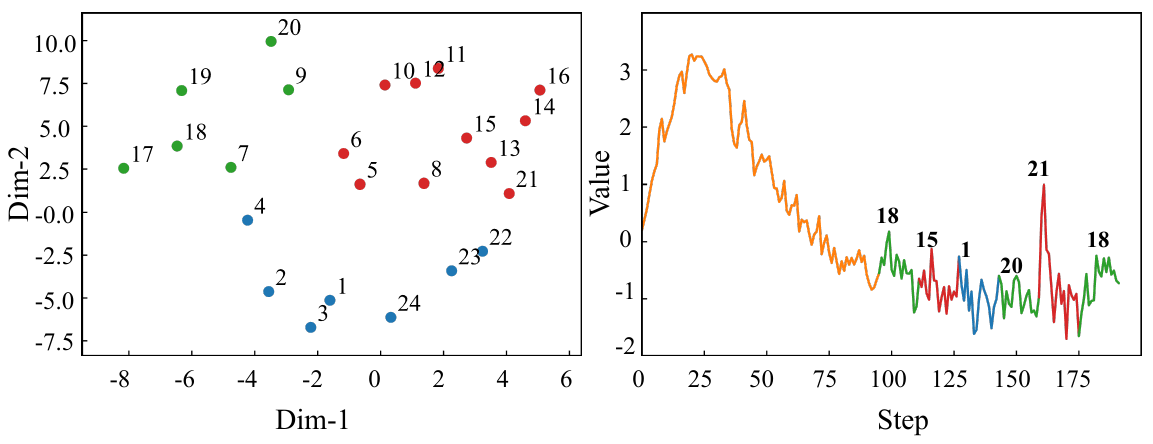}
    \caption{Illustration of codebook clustering in the latent space and dynamic reconstruction.}
    \label{fig:reconstruction}
\end{figure}

\paragraph{Qualitative Analysis of Tokenization. }
As shown in Figures \ref{fig:codebook_heatmap} and \ref{fig:reconstruction}, we conduct a comprehensive evaluation of the proposed discretization module on the Nature dataset from three complementary perspectives.
The token usage heatmap (Figure \ref{fig:codebook_heatmap}) indicates that all 64 tokens are actively and consistently utilized across samples, demonstrating effective mitigation of codebook collapse and a strong capacity to capture diverse and heterogeneous temporal structures.
The codebook clustering visualization (Figure \ref{fig:reconstruction}, left) further reveals that tokens organize into coherent and well-separated groups in the latent space, suggesting that the learned discrete vocabulary preserves meaningful structural relationships among different temporal patterns.
Moreover, the dynamic reconstruction results (Figure \ref{fig:reconstruction}, right) highlight the tokenizer’s context-adaptive decoding behavior: the same token id (e.g., ID = 18) can generate distinct decoded segments under different contextual conditions, thereby enabling faithful and flexible alignment with the original time series.
Overall, these empirical findings confirm that the proposed discretization process learns a diverse, semantically organized, and structurally meaningful vocabulary, while effectively supporting context-aware and adaptive decoding for time series forecasting.

\section {Conclusion}
We proposed TokenCast, a context-aware TSF framework based on a pretrained LLM. This approach first converts a continuous time series into discrete tokens. Leveraging a pretrained LLM, it aligns the temporal and contextual tokens through an autoregressive objective, achieving unified modeling of both modalities. The model is then further fine-tuned to generate future token sequences.
We evaluate TokenCast on multiple real-world datasets rich in contextual information. Experimental results demonstrate that TokenCast achieves superior accuracy. We also conduct comprehensive ablation experiments and qualitative analysis to validate the framework's adaptability and flexibility for symbolic, LLM-driven TSF. Looking ahead, we believe that leveraging language as a symbolic intermediary will have the potential to advance TSF towards a multimodal and multi-task level.

\section*{Acknowledgments}
This research was supported by grants from the National Natural Science Foundation of China (No. 62502486, 62376193), the grants of the Provincial Natural Science Foundation of Anhui Province (No. 2408085QF193), Guangdong S\&T Programme (No. 2025B0101120004), USTC Research Funds of the Double First-Class Initiative (No. YD2150002501), the Fundamental Research Funds for the Central Universities of China (No. WK2150110032).

\appendix

\bibliographystyle{named}
\bibliography{ijcai26}

\end{document}